\documentclass{edm_article}

\usepackage{dirtytalk} 
\usepackage{amsmath} 
\usepackage{array}
\usepackage{dsfont}
\usepackage{xcolor}
\usepackage{bbm}

\usepackage{url}

\usepackage[scr]{rsfso}

\usepackage{mathtools}

\usepackage{siunitx}
\usepackage{booktabs}
\usepackage[version=4]{mhchem}
\def\x{\boldsymbol{\mathrm{x}}}

\usepackage{hyperref}
\hypersetup{hidelinks,,pdfpagelabels=false}

\begin{document}

\title{Towards Scalable Adaptive Learning with Graph Neural Networks and Reinforcement Learning}




\renewcommand{\affaddr}[1]{\ensuremath{\textrm{#1}}}
\renewcommand{\email}[1]{\href{mailto:#1}{\texttt{#1}}}
\numberofauthors{1}
\author{
    \alignauthor
    Jean Vassoyan\textsuperscript{\textsf{1,2}} \qquad
    Jill-Jênn Vie\textsuperscript{\textsf{3}} \qquad
    Pirmin Lemberger\textsuperscript{\textsf{2}}\\[5mm]
    \begin{tabular}{@{}l@{}}
      \affaddr{\textsuperscript{\textsf{1}} Université Paris-Saclay, CNRS, ENS Paris-Saclay, Centre Borelli,} \email{jean.vassoyan@ens-paris-saclay.fr}
    \end{tabular}\\[1mm]
    \begin{tabular}{@{}l@{}}
      \affaddr{\textsuperscript{\textsf{2}} onepoint,} \email{p.lemberger@groupeonepoint.com}
    \end{tabular}\\[1mm]
    \begin{tabular}{@{}l@{}}
      \affaddr{\textsuperscript{\textsf{3}} Inria Saclay, SODA,} \email{jill-jenn.vie@inria.fr}
    \end{tabular}
}

\maketitle

\begin{abstract}

Adaptive learning is an area of educational technology that consists in delivering personalized learning experiences to address the unique needs of each learner.
An important subfield of adaptive learning is learning path personalization: it aims at designing systems that recommend sequences of educational activities to maximize students' learning outcomes.
Many machine learning approaches have already demonstrated significant results in a variety of contexts related to learning path personalization.
However, most of them were designed for very specific settings and are not very reusable.
This is accentuated by the fact that they often rely on non-scalable models, which are unable to integrate new elements after being trained on a specific set of educational resources.
In this paper, we introduce a flexible and scalable approach towards the problem of learning path personalization, which we formalize as a reinforcement learning problem.
Our model is a sequential recommender system based on a graph neural network, which we evaluate on a population of simulated learners.
Our results demonstrate that it can learn to make good recommendations in the small-data regime.
\end{abstract}


\section{Introduction}

Adaptive learning is an area of educational technology that focuses on addressing the unique needs, abilities, and inte-rests of each individual student.
This field emerged in the 1980s with the introduction of the first \textit{Intelligent Tutoring Systems} (ITS) and experienced major expansion in the 1990s.
As described by T. Murray in \cite{murray1999authoring}, an ITS usually consists of four components: a \textit{domain model}, a \textit{student model}, an \textit{instructional model} and a \textit{user interface model}.
As we address the problem from an algorithmic point of view, we only focus on the first three models.
The domain model is a representation of the knowledge to be taught; it often serves as a basis for the student model.
The student model provides a characterization of each learner that allows to assess their knowledge and skills and anticipate their behavior.
The instructional model takes the domain and student models as input to select strategies that will help each user achieve their learning objectives.
This general structure allows ITSs to achieve many purposes (recommending exercises, providing feedback, facilitating memorization, etc.) while optimizing a variety of metrics (learning gains, engagement, speed of learning, etc.).

In this paper, we address the problem of learning path personalization with optimization of learning gains.
This means that we look for a sequential recommender system that can provide each student with the right content at the right time (according to their past activity), in order to maximize their overall learning gains.

Towards this goal, \say{standard} approaches often require significant structuring of the domain model.
This step is usually assisted by experts: they may be mobilized to tag educational resources, set up prerequisite relationships, draw up skill tables, etc.
One example of such structuring is  the Q-matrix \cite{barnes2005q} which maps knowledge components (KC) to exercises. 
These expert-based approaches present some serious practical limitations.
First, they make it quite cumbersome to create resource sets, since each resource has to be properly tagged (sometimes with an extensive set of metadata).
They also lead to poorly reusable recommender systems, since prerequisite relationships and skills maps are usually tailored to specific resource sets.
This low reusability problem is often exacerbated by the modeling of resources/skills/KC as one-hot encodings \cite{bassen2020reinforcement, piech2015deep} which tie the model to a maximum number of resources/skills/KC it can handle.
As a result, these approaches produce models that are not suitable for transfer learning.
Our approach, on the other hand, is based on a graph neural network, which structure makes it possible to process data in a much more flexible way.

Our contributions in this paper are threefold.
    First, we introduce a new setting for learning path personalization and formalize it as a model-free reinforcement learning (RL) problem.
    Second, we present a novel RL policy that can leverage educational resource content and users' feedback to make recommendations that improve learning gains.
    The proposed model has the advantage of being inherently scalable, reusable, and independent of any expert tagging.
    Third we evaluate our model on 6 semi-synthetic environments composed of real-world educational resources and simulated learners.
    The results demonstrate that it can learn to make good recommendations from few interactions with learners, thereby significantly outperforming the uniform random policy.

The rest of the paper is organized as follows.
In Section \ref{section:related_works}, we relate our paper to prior research.
In Section \ref{section:problem_formulation}, we describe our setting, the assumptions we make and the problem we attempt to solve, which we formalize as a reinforcement learning problem.
In Section \ref{section:rl_agent}, we present our novel RL policy.
In Section \ref{section:experiments}, we describe our experimental setting and discuss our results. In Section \ref{section:future_works} we address some limitations of our model and propose a few directions for future work.
We finally conclude in Section \ref{section:conclusion}.

\section{Related Work}\label{section:related_works}

In recent years, several works have used reinforcement learning to address the problem of learning path personalization.
Most of these RL approaches are model-based, as they rely on a predefined student model to simulate student trajectories. 
However, no student model is completely accurate, and the learned instructional policies may overfit to the student model.
Doroudi et al. \cite{doroudi2017robust} have attempted to learn policies that provide a better reward no matter the student model chosen (i.e. robust policies).
Azhar et al. \cite{azhar2022optimizing} proposed a method to gradually refine the student model by adding features that maximize the reward.

Reward functions usually involve learning gains.
Subramanian and Mostow \cite{subramanian2021deep} defined learning gains as average difference between posterior and prior latent knowledge.
Lan and Baraniuk \cite{lan2016contextual} proposed to learn a policy for selecting learning actions so that the grade on the next exam is maximized.
Clement et al. \cite{clement2015multi} attempted to optimize an increase in success rate in recent time steps, they used an $\varepsilon$-greedy approach.
Doroudi et al. \cite{doroudi2019s} conducted a thorough review of the different reward functions used in instructional policies.

The closest to our setting is probably the approach proposed by Bassen et al. \cite{bassen2020reinforcement} which, like ours, does not rely on expert pre-labeling of educational resources.
However, in the absence of compensation for this lack of information, their reinforcement learning algorithm requires a substantial number of learners to converge to an effective policy: about 1000 learners for a corpus of 12 educational resources.
Moreover, in their framework, educational activities were represented as one-hot encodings and passed to the policy via a fixed-size vector.
Therefore, this approach does not allow to work with an evolving corpus of educational resources (which is the case for most \textit{e-learning} platforms) nor to reuse the model on another set, unless it is completely re-trained.

In contrast, our approach leverages information from resource keywords which allows to achieve convergence in a relatively small number of episodes, while maintaining a high level of flexibility.
This keyword-based approach was inspired by the work of Gasparetti et al. \cite{gasparetti2015exploiting, gasparetti2018prerequisites}.
Although the authors did not directly address the problem of learning path personalization, they outlined a method of feature extraction from textual resources that proved to be very successful in predicting prerequisite relationships.



\section{Problem formulation}\label{section:problem_formulation}

\subsection{Description of the setting}\label{section:description}
Consider an \textit{e-learning} platform with a collection of educational resources which have been designed to cover a specific topic, for example \say{an introduction to machine learning}.
Consider a population $\mathscr{P}$ of learners to be trained on this topic.
The goal of learning path personalization is to be able to recommend a sequence of educational resources to each learner so as to maximize his overall \textit{learning gains}.
Therefore the resulting machine learning problem can be expressed in the following terms: given a large enough sample ${U}$ of users from $\mathscr{P}$, how can we train a machine learning model to make recommendations to users from $U$ so as to generalize to the whole population?

In this paper, we work at the scale of short learning paths ($\sim$ 1 hour), which means that each learning session only consists of a few interactions between the learner and the ITS.
One advantage of this setting is that it reduces the effects of memory loss: we assume that when a learner visits a new resource, what he learned from the previous ones is still in his working memory.

\begin{figure}
    \centering
    \includegraphics[width=\linewidth]{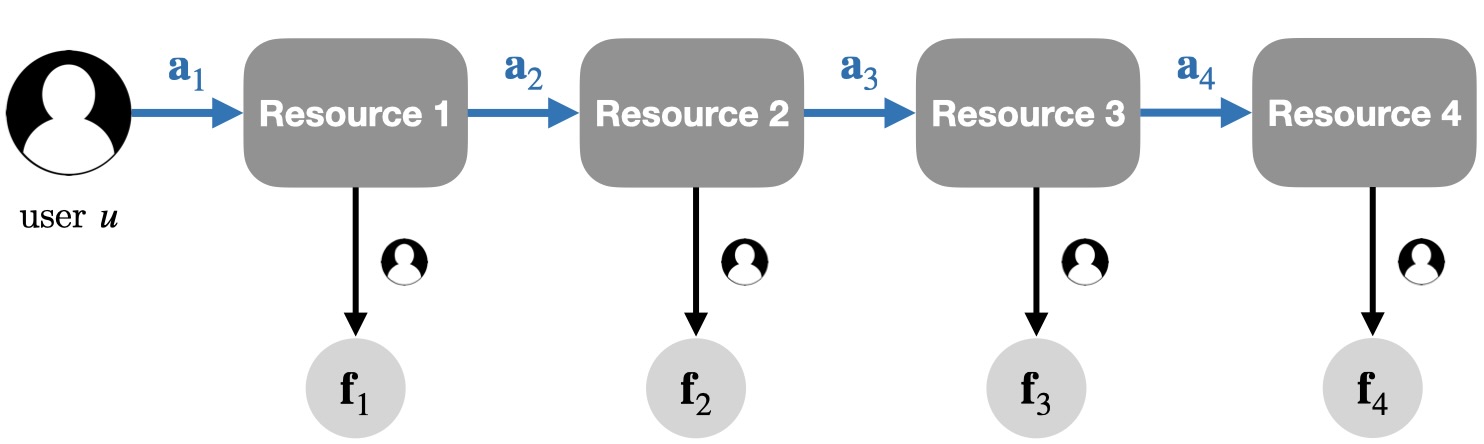}
    \caption{A view of a learning session. In this example, the session length is $T = 4$.
    Actions $\mathbf{a}_1,\mathbf{a}_2,\mathbf{a}_3,\mathbf{a}_4$ are the recommendations of the ITS.
    $\mathbf{f}_1,\mathbf{f}_2,\mathbf{f}_3,\mathbf{f}_4$ are the feedback signals returned by the user.}
    \label{fig:user_session}
    \Description{A user goes through several documents and gives feedback to each of them.}
\end{figure}

\def\feq{f_\circ}
We first make a few assumptions about the learning sessions:%
\begin{itemize}
    \item[$(a_1)$] Each learner follows one learning path of equal length (i.e. same number of resources).
    The purpose of this assumption is primarily to simplify the notations as it can be easily relaxed without making major modifications to the model.
    \item[$(a_2)$] There is no interaction between the learner and the external world (no communication, no access to external resources). This makes it possible to work in the closed system \{learner + ITS\}. While incorrect in most cases, this assumption may be more reasonable in our setting than in a multi-day learning context.
    \item[$(a_3)$] We assume the existence of a feedback signal that provides information about user understanding of each resource.
    This signal can take three values:
    \begin{itemize}
        \item $(f_{<})$: the user did not understand the resource
        \item $(f_>)$: the user understood, but found it too easy
        \item $(\feq)$: the resource was at the right level.
    \end{itemize}
    In practice, such feedback can be obtained from self-assessment or more sophisticated test, and should be associated with an error margin to account for its imprecision.
    Nevertheless, in this study, we assume that each feedback is perfectly accurate.
\end{itemize}%
A view of such a learning session is provided in Figure \ref{fig:user_session}.

To further simplify the problem, we also adopt a few simplifying assumptions about the educational resources:
\begin{itemize}
    \item[$(a_4)$] They are purely textual resources, written in natural language. We indeed consider that most educational formats can be easily transcribed into text (transcript of a video, legend of a diagram, caption of an image etc.).
    \item[$(a_5)$] They are \textit{self-contained}, which means that they can be considered independently. This implies for example that they do not explicitly refer to each other.
    Although quite strong, this assumption is essential to prevent mandatory dependencies and foster diversity of learning paths.
    \item[$(a_6)$]
    Each resource explains one or few concepts and has equivalent \say{educational value}. This involves that each resource carries the same \say{amount} of knowledge.
\end{itemize}%
Some examples of educational resources that satisfy these requirements are provided in Figure \ref{fig:resources} of the Appendix.

Our goal with this work is to design a machine learning algorithm that can leverage learners' feedback to text-based educational resources to model their understanding of each concept, anticipate their reactions, and recommend resources that maximize their overall learning gains.

Since most \textit{e-learning} platforms are in constant evolution, our goal is not only to solve this problem but to do it in a flexible and scalable way.
This means that the model should not require full retraining when new resources are added to (or removed from) the platform.
Actually, it should be able to extrapolate to new resources what it learned from previous interactions.
This suggests that the number of parameters of our model should not depend on the size of the corpus.

\subsection{Formalization}\label{section:formalization}
In this section, we formalize the problem described above as a reinforcement learning problem.
We use the terms \say{user} and \say{learner} interchangeably to refer to any individual from the sample $U$.
Similarly, we refer to an educational resource with the terms \say{document} or \say{resource}.

In the following, we denote:
    $T$ the length of each learning session (identical for each user),
    $\mathscr{D}$ the corpus of documents,
    $d$ a document from $\mathscr{D}$,
    $u$ a user (or learner) from the sample ${U}$,
    $\mathbf{f}_{d}$ the feedback given by a learner on document $d$.

The sequential recommendation problem defined above can be easily expressed as a reinforcement learning problem \break where: the agent is the recommender system, the environment is the population $\mathscr{P}$ of students and each episode is a learning path.
This problem can be formulated as a partial-ly observable Markov decision process \mbox{($\mathcal{S}$, $\mathcal{A}$, $\mathcal{O}$, $\mathcal{T}$, $\mathcal{R}$, $\mathcal{Z}$)} where $\mathcal{S}$ is the state space, $\mathcal{A}$ is the action space, $\mathcal{O}$ is the observation space, $\mathcal{T}:\mathcal{S}\times\mathcal{A}\times\mathcal{S} \rightarrow[0,1] $ defines the conditional transition probabilities, $\mathcal{R}: \mathcal{S} \times \mathcal{A} \rightarrow \mathbb{R}$ is the reward function and $\mathcal{Z}: \mathcal{S} \times \mathcal{A} \times \mathcal{O} \rightarrow [0,1]$ is the observation function.
More precisely, in our setting:
\begin{itemize}
    \item $\mathbf{s}_t\in \mathcal{S}$ is the (unknown) knowledge state of the learner at step $t$;
    \item $\mathbf{a}_t\in \mathcal{A}$ is the document selected by the recommender system at step $t$;
    we can write $\mathbf{a}_t=\mathbf{d}_t$;
    \item $\mathbf{o}_t\in \mathcal{O}$ is the observation made at step $t$, which is a tuple of the selected document and the returned feedback: $\mathbf{o}_t =  (\mathbf{d}_t, \mathbf{f}_t)$;
    \item $\mathcal{T}(\mathbf{s},\mathbf{a},\mathbf{s}^{\prime})=\mathbb{P}(\mathbf{s}_{t+1}=\mathbf{s}^{\prime}\mid \mathbf{s}_t=\mathbf{s},\mathbf{a}_t=\mathbf{a})$ is unknown, as it represents the impact of selecting document $\mathbf{a}$ on learner's state $\mathbf{s}_t$;
    \item $\mathcal{Z}(\mathbf{s}, \mathbf{a}, \mathbf{o})=\mathbb{P}(\mathbf{o}_{t+1}=\mathbf{o}\mid \mathbf{s}_{t+1}=\mathbf{s}, \mathbf{a}_t=\mathbf{a})$ is also unknown and represents the probability of observing $\mathbf{o}$ in state $\mathbf{s}$ after choosing document $\mathbf{a}$;
    \item $\mathcal{R}(\mathbf{s}_t,\mathbf{a}_t)$ is the learning gain of the user at step $t$, which we define as follows:
    \begin{eqnarray}\label{eq:reward}
        \mathcal{R}(\mathbf{s}_t,\mathbf{a}_t) = \mathbbm{1}_{\{\mathbf{f}_t = \feq\}}.
    \end{eqnarray}
    We indeed consider that only feedback $\feq$ corresponds to an effective learning gain.
    We denote $\mathcal{R}(\mathbf{s}_t,\mathbf{a}_t)=\mathbf{r}_t$ in the following.
\end{itemize}
To solve this problem, we need to find a policy \mbox{$\pi :\mathcal{O}\rightarrow \mathcal{A}$} that maximizes the expected return over each episode $\eta$:
\begin{equation}
    \pi^* = \underset{\pi}{\arg \max }\ \mathbb{E}_{\eta\sim\pi} \left[\  \sum_{t=1}^{T} \mathbf{r}_t \ \right].
\end{equation}
\section{Our RL model}\label{section:rl_agent}



\begin{figure}[t]
    \centering
    \includegraphics[width=0.9\linewidth]{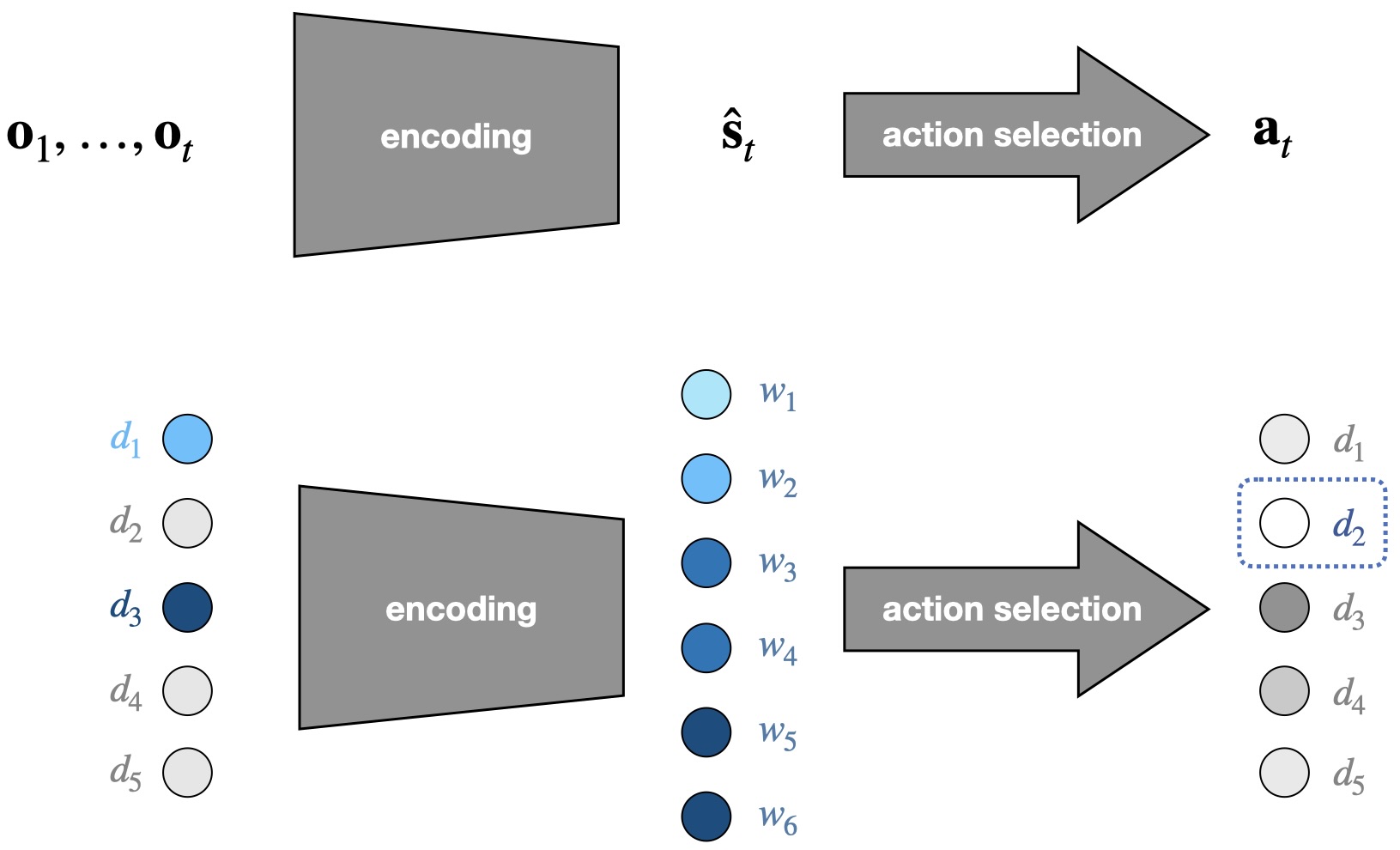}
    \caption{Up, a view of common policy architecture to solve POMDP. Down, this architecture applied to our setting.}
    \Description{Up, a latent state is encoded from observations and the policy chooses actions from this latent state. Down, from observed feedback, latent representation is estimated at the keyword level from which new documents are selected by our policy}
    \label{fig:latent_space}
\end{figure}

A common approach to solve partially observable Markov decision processes (POMDP) is to leverage information from past observations $\mathbf{o}_1, \dots, \mathbf{o}_t$ to build an estimation of $\mathbf{s}_t$ which is then used to select the next action (illustrated in Figure \ref{fig:latent_space}).
This boils down to encoding these observations into a latent space $\mathcal{S}$.
In our setting, this latent space contains all possible knowledge states for the learner, which is why we call it \textit{knowledge space} in the following.

\subsection{Knowledge space}

\begin{figure*}[t]
    \includegraphics[width=\textwidth]{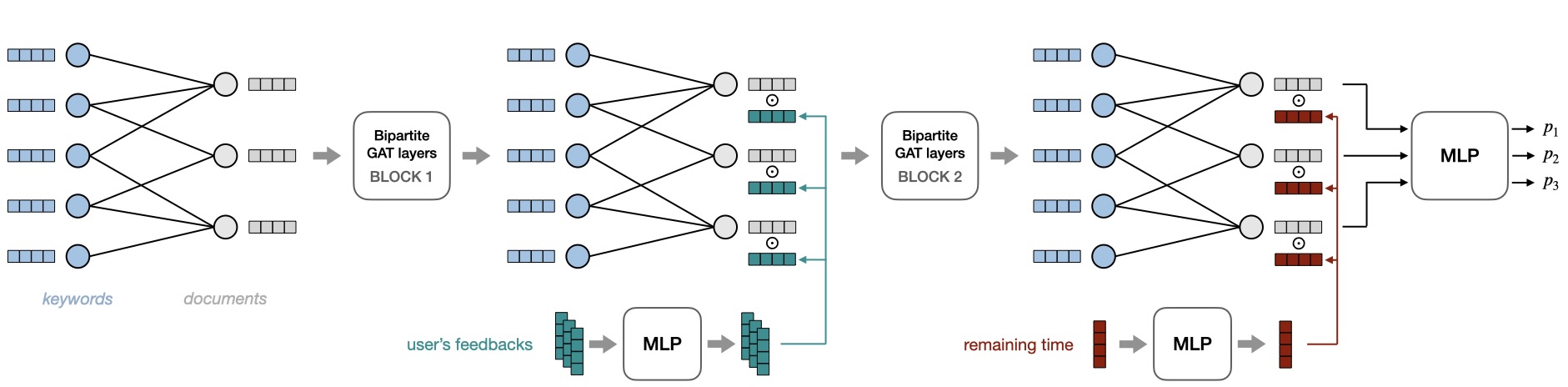}
    \caption{The architecture of our policy network on a 3-document corpus}
    \label{fig:full_model}
    \Description{Diagram of the architecture of our neural network.}
\end{figure*}

While more compact than the observation space, the knowledge space should be informative enough to convey a relevant approximation of learner's knowledge.

We decided to structure this representation with the keywords of the corpus, denoted $(w_1, \dots, w_M)$.
We define a keyword as a word or group of words that refers to a technical concept closely related to the subject of the corpus.
Some examples of keywords extracted from educational resources are provided in the Appendix.
The keywords carry information about the concepts addressed by the documents and are therefore a good approximation of their pedagogical content.
That is why we modeled the knowledge state of each learner as a collection of vectors $(\mathbf{w}_1, \dots, \mathbf{w}_M)$ which represents his \say{understanding} of each keyword.
We indeed consider that a keyword can be understood in multiple ways depending on the context in which it occurs, and a multidimensional vector can be a convenient way to capture this plurality.
This is illustrated in Figure \ref{fig:latent_space}.
From this perspective, the knowledge space $\mathcal{S}$ can be defined as \mbox{$\mathcal{S} \vcentcolon = \underbrace{\mathbb{R}^K\times\dots\times\mathbb{R}^K}_{ M}$}.

Note that we do not consider keyword extraction as a task requiring expert knowledge since it can be done by any creator of educational content and involves fewer skills than defining the knowledge components of a course.
Moreover, it is mainly a pattern-matching task that can be automated through a keyword extraction algorithm \cite{tagme, rake, yake}.


\subsection{Policy}

Following the previous considerations, the policy $\pi_{\theta}$ should take a collection of observations $\mathbf{o}_1,\dots,\mathbf{o}_t$ as input, encode it into the latent space $\mathcal{S}$ and return a recommendation for the next document $\mathbf{d}_t$.
We emphasize that this function should also meet the aforementioned flexibility and scalability requirements. 

A natural way to model the relationship between documents and keywords is to build a bipartite graph $\mathcal{G} = (\mathcal{V}_{\scriptscriptstyle\mathcal{D}}, \mathcal{V}_{\scriptscriptstyle \mathcal{W}}, \mathcal{E})$, where
$\mathcal{V}_{\scriptscriptstyle\mathcal{D}}$ is the set of \textit{document} nodes, $\mathcal{V}_{\scriptscriptstyle\mathcal{W}}$ is the set of \textit{keyword} nodes and $\mathcal{E}$ is the set of edges, with $({v}_d,{v}_w)\in\mathcal{E}$ if the document $d$ contains the word $w$.

We chose to use a graph neural network (GNN) as a policy.
GNNs are quite convenient for this task as they allow to enrich node features with information about their extensive neighborhood, through message-passing.
Therefore, documents (respectively keywords) that share a large number of keywords (respectively documents) will also have similar embeddings.
This allows to build keyword embeddings that contain information about feedback from neighboring documents ($ \mathbf{o}_1,\dots,\mathbf{o}_t \rightarrow \hat{\mathbf{s}}_t$).
Message-passing can also be used the other way around, from keywords to documents, to build embeddings that inform about the relevance of each document according to the estimated knowledge state ($\hat{\mathbf{s}}_t \rightarrow \mathbf{a}_t$).
Another significant advantage of GNNs is that their number of parameters does not depend on the size and structure of the graph, which makes them highly flexible and scalable.

Multiple options are possible for the initial node features.
For keyword nodes, pre-trained word embeddings are a natural choice.
As for the document nodes, a simple null vector is sufficient.
However one may choose to include extra information about the documents if it is available (type of document, format, length etc.).
We denote as ${(\mathbf{x}_w)}_{w\in\mathcal{V}_{\scriptscriptstyle\mathcal{W}}}$ and $(\mathbf{x}_{d})_{d\in\mathcal{V}_{\scriptscriptstyle\mathcal{D}}}$ the initial feature vectors of keyword and document nodes.

In our model, we adapted a version of GAT (graph attention networks) \cite{velivckovic2017graph} to the heterogeneity of our bipartite graph:
\begin{align} 
    \label{eq:kw2doc}
    \forall d\in\mathcal{V}_{\scriptscriptstyle\mathcal{D}},\; \mathbf{h}_d^{(\ell+1)}&=\sigma\left(\sum_{w \in \mathcal{N}(d)} \alpha_{d w}^{(\ell)} W_{D}^{(\ell)} \mathbf{h}_w^{(\ell)} +B_{D}^{(\ell)}\right)\\
    \label{eq:doc2kw}
    \forall w\in\mathcal{V}_{\scriptscriptstyle\mathcal{W}}, \mathbf{h}_w^{(\ell+1)}&=\sigma\left(\sum_{d \in \mathcal{N}(w)} \alpha_{w d}^{(\ell)} W_{W}^{(\ell)} \mathbf{h}_d^{(\ell+1)} +B_{W}^{(\ell)}\right)
\end{align}
$\mathbf{h}^{(\ell)}_d\in\mathbb{R}^{K}$ is the embedding of node $d$ at $\ell$th layer, with $\mathbf{h}^{(0)}_d=\mathbf{x}_d$.
$\mathcal{N}(d)$ is the set of neighbors of node $d$ in the graph.
$\alpha_{d w}^{(\ell)}$ is a \textit{self-attention} coefficient, detailed in the Appendix.
$\sigma(\cdot)$ is the ReLU activation function (rectified linear unit).
$W_W^{(\ell)}$, $W_D^{(\ell)}$, $B_W^{(\ell)}$ and $B_D^{(\ell)}$ are trainable parameters.
This back-and-forth mechanism between documents and keywords allows to learn distinct filters for each node type (document or keyword), effectively addressing the graph's heterogeneity.
In the following, we refer to equations (\ref{eq:kw2doc}) and (\ref{eq:doc2kw}) as \textit{bipartite GAT layers} and denote them ($\text{KW} \overset{\text{(\ref{eq:kw2doc}})}{\longrightarrow} \text{DOC}$) and ($\text{DOC} \overset{\text{(\ref{eq:doc2kw}})}{\longrightarrow} \text{KW}$).
Note that they can be chained one after the other.

We define our first block of bipartite GAT layers as follows:
\begin{equation}
    \text{BLOCK}1 \ =\  \text{KW} \overset{\text{(\ref{eq:kw2doc}})}{\longrightarrow} \text{DOC} \overset{\text{(\ref{eq:doc2kw}})}{\longrightarrow} \text{KW} \overset{\text{(\ref{eq:kw2doc}})}{\longrightarrow} \text{DOC}.
\end{equation}
After this block, document embeddings $(\mathbf{h}_d^{(2)})_{d\in\mathcal{V}_{\scriptscriptstyle\mathcal{D}}}$ contain information about keywords from their extended neighborhood.
Using a Hadamard product, we enrich these embeddings with user feedback:
\begin{equation}
    \mathbf{h}_d^{(\varphi)} = \mathbf{h}_d^{(2)} \odot \textsf{MLP}_{K_d \to K}(\mathbf{f}_d)
\end{equation}
$\mathbf{h}_d^{(2)}$ and $\mathbf{h}_d^{(\varphi)}$ are the embeddings of document $d$ before and after adding the feedback.
${\mathbf{f}_d}$ is an encoding of user's feedback on document $d$,
which is passed through a multilayer perceptron (MLP).
We use a \say{not visited} feedback for the documents that the learner has not yet visited.

After doing this operation on each document node, we apply another block of bipartite GAT layers:
\begin{equation}\label{eq:block2}
    \text{BLOCK}2 \ =\ \text{DOC} \overset{\text{(\ref{eq:doc2kw})}}{\longrightarrow} \text{KW} \overset{\text{(\ref{eq:kw2doc})}}{\longrightarrow} \text{DOC}.
\end{equation}
Operation $\text{(\ref{eq:doc2kw})}$ allows to enrich keyword embeddings with feedback from neighboring documents, which carry information about user's understanding.
We consider these embeddings as a good approximation of learner's knowledge state, which is why we define $\hat{\mathbf{s}}_t \vcentcolon = (\mathbf{h}_w^{(2)})_{w\in\mathcal{V}_{\scriptscriptstyle\mathcal{W}}}$.
The final GAT layer $\text{(\ref{eq:kw2doc})}$ maps $\hat{\mathbf{s}}_t$ to documents for the next recommendation.

Before assigning probabilities to each document in the final step, we enrich document embeddings by incorporating information about the remaining time in the session, which, as we observed, slightly improved the performance of the model:
\begin{equation}
    \mathbf{h}_d^{(\tau)} = \mathbf{h}_d^{(3)} \odot \textsf{MLP}_{K_\tau \to K}(\Delta_t)
\end{equation}
$\mathbf{h}_d^{(3)}$ and $\mathbf{h}_d^{(\tau)}$ are the embeddings of document $d$ before and after adding the remaining time.
$\Delta_t = T-t$ is an encoding of the remaining time (or remaining steps) at step $t$.

Eventually, the embeddings $\mathbf{h}_d^{(\tau)}$ are passed through an MLP to assign a score to each document.
These scores are converted into probabilities via a softmax over all document nodes (further details in the Appendix):
\begin{equation}
    \pi_{\theta}\left(d\mid \mathbf{o}_1,\dots,\mathbf{o}_t\right) = \underset{\mathcal{V}_{\scriptscriptstyle\mathcal{D}}}{\operatorname{softmax}} \left( \textsf{MLP}_{K \to 1}(\mathbf{h}_d^{(\tau)}) \right).
\end{equation}
The full architecture of the policy is illustrated in Figure \ref{fig:full_model}.


\subsection{RL Algorithm}

As our policy selects the next action directly from observations, it belongs to the \textit{policy-based} reinforcement learning paradigm, especially the \textit{policy gradient} methods.
The latter make it possible to maximize the expected return by optimizing directly the parameters of $\pi_{\theta}$ through gradient descent.
We chose the \texttt{REINFORCE} algorithm \cite{sutton1999policy} for its simplicity.
At the end of each episode, $\pi_{\theta}$ is updated as follows:
\begin{equation}
    \forall t\in[1,T],\quad \theta \leftarrow \theta+\lambda \nabla_\theta \log \pi_\theta\left(s_t, a_t\right) v_t
\end{equation}
with $\lambda$ the learning rate and $v_t = \sum_{t^\prime=t}^T \gamma^{t^\prime-t}r_{t^\prime}$ the return of the episode from step $t$.

Note that we could learn our policy using more sophisticated RL algorithms like actor-critic, which usually has lower variance.
However, it is likely that the current architecture would provide a poor state value function as it only operates at the scale of node neighborhoods and does not have a \say{global} view of the graph.
Some changes in this architecture might nevertheless be done to process information at a larger scale, as discussed in Section \ref{section:future_works}.

\section{Experiments}\label{section:experiments}

Given the complexity of conducting mass experiments on real learners, we chose to evaluate our model in an environment made up of semi-synthetic data.
Our implementation is written in Python and is available on GitHub\footnote{\href{https://github.com/jvasso/graph-rl4adaptive-learning}{https://github.com/jvasso/graph-rl4adaptive-learning}}.
We also provided the hyperparameters of our model in Table \ref{table:hyperparameters} of the Appendix.
\subsection{Experimental setting}
\subsubsection*{Linear corpus}

\begin{table}[t]
    \centering
    \medskip
    \begin{tabular}{ccccc}
        \toprule
        corpus   & \# doc & \# kw & \# edges & diameter\\
        \midrule
        Corpus 1 & 33 & 68 & 154 & 10\\
        Corpus 2 & 11 & 31 & 62  & 6\\
        Corpus 3 & 19 & 39 & 83  & 8\\
        Corpus 4 & 28 & 55 & 113 & 8\\
        Corpus 5 & 18 & 41 & 66  & $\infty$\\
        Corpus 6 & 20 & 45 & 143 & 6\\
        \bottomrule
    \end{tabular}
    \caption{Key statistics of each corpus}
    \label{table:corpora}
\end{table}

We introduce what we call a \say{linear} corpus.
Starting from a regular course divided into sections and subsections, we treat each subsection as one document.
The corpus resulting from this decomposition is \say{linear}, in the sense that it was designed to be followed in a single, pre-defined order, which is identical for each learner.
Therefore, it leaves practically no room for personalization.
Six corpora were constructed this way: three about data science (1-3) and three about programming (4-6).
They were all built from courses taken from a popular \textit{e-learning} platform.

For the purpose of our experiments, we have chosen to tag keywords \say{by hand} to avoid introducing any noise in the results.
Our methodology was quite simple: for each document, we collected keywords referring to technical concepts related to the topic of the course.
Table \ref{table:corpora} presents some key statistics about each corpus and their associated bipartite graphs.
Note that the graph of corpus 5 is disconnected: indeed, one of its documents only contains keywords that do not appear in any other document.
Despite significantly complicating the task for a diffusion model like ours, we have chosen to keep this corpus for our experiments.

\subsubsection*{Simulated learners}\label{subsection:simulation}

Since each corpus has been designed to be explored in a single pre-defined order, we assume that the only way to understand it is to follow this order scrupulously.
Therefore we have decided to simulate the behavior of learners in this very simple way: as long as the policy recommends documents in the right order, the learner returns the feedback $(\feq)$.
Conversely, each time the algorithm recommends a document too early or too late, the learner returns the feedback $(f_{<})$ or $(f_{>})$.
A detailed example is given in the Appendix.

Since our simulated learners have a straightforward behavior, the purpose of this experiment is not to evaluate the personalization or generalization capabilities of our model, but to assess its ability to grasp the structure of a corpus, by finding its original order in a reasonable number of episodes (i.e. a few learners).
While trivial at first glance, this task can be quite difficult for an RL agent in the small-data regime.
Besides, each corpus contains some parts that are independent of each other which suggests that in practice, multiple learning trajectories might be understandable to real learners.
From this perspective, the \say{strict} feedback of our simulated learners can distort the real nature of the relationships between resources and make the task more difficult for our recommender system.

\subsubsection*{Policy}

In our experiments, we compared 3 different policies.
The first one is the uniform random policy.
The second one is our policy with one-hot-encodings as keyword features.
The third one is our policy with \textsf{Wikipedia2Vec} embeddings \cite{wikipedia2vec} as keyword features.
\textsf{Wikipedia2Vec} embeddings are quite suitable for our task as they contain encyclopedic information about the relationship between words and entities.
They were derived from a skip-gram model trained on a triple objective, which is detailed in the Appendix.
We used null vectors as document features for each policy.

\subsubsection*{Training}

In each experiment, the maximum achievable return is equal to the size of the corpus.
We set the horizon $T$ to the size of the corpus to make sure that only an optimal policy (i.e. one that makes no \say{mistake}) can reach this return.
In this setting, the return of the random policy follows a binomial distribution with parameters ($T$, $\frac{1}{T}$).
Therefore its expected return is 1 for each episode.
We also set the discount factor $\gamma=0$ during training because in this very specific setting, the best action at each step $t$ can be learned from immediate reward.
We trained our model from scratch over 50 episodes ($\sim$ 50 students) for each corpus, with a constant learning rate.


\subsection{Results}\label{section:results}

\begin{table}[b]
    \centering
    \medskip
    \begin{tabular}{ccc}
        \toprule
        Corpus & \textsf{Wikipedia2Vec} & {One-hot encodings}  \\
        \midrule
        Corpus 1 & $\mathbf{16.48 \pm 2.66}$ & $13.36 \pm 1.74$  \\
        Corpus 2 & $\mathbf{10.84 \pm 0.14}$ & $10.28 \pm 0.37$  \\
        Corpus 3 & $\mathbf{14.40 \pm 1.31}$ & $11.68 \pm 1.13$  \\
        Corpus 4 & $\mathbf{15.16 \pm 0.90}$ & $12.52 \pm 0.98$  \\
        Corpus 5 &  $\mathbf{9.80 \pm 2.13}$ & $7.56  \pm 1.83$  \\
        Corpus 6 & $\mathbf{11.24 \pm 1.52}$ & $8.24  \pm 0.84$  \\
        \bottomrule
    \end{tabular}
    \caption{Comparison between episodic returns when using \textsf{Wikipedia2Vec} and one-hot encodings as keyword features}
    \label{table:results}
\end{table}

Since the \texttt{REINFORCE} algorithm has quite a high variance, we averaged each episodic return over 25 random seeds.
The resulting learning curves are shown in Figure \ref{fig:results_curves} of the Appendix and the last episodic returns (measured at 50\textsuperscript{th} episode) are reported in Table \ref{table:results}.

From these curves, one can notice that despite the small-data regime and the choice of a sub-optimal RL algorithm (the \texttt{REINFORCE} algorithm is known to be quite unstable and sample-inefficient), our agent succeeded in recovering a significant part of the original order of each corpus.
Most of the time, it achieved average return over 10 whereas the random policy was stuck in an expected return of 1.

Best performance was achieved on Corpus 2.
Indeed, it is the only one for which our model managed to reach the maximum achievable return most of the time.
This may be partly due to the small number of documents in this corpus.
However, we stress that the number of documents alone is not a sufficient feature to account for the variability of the results.
For instance, corpora 3 and 6 have a nearly similar number of documents, but our model performed very differently on these two corpora.
Moreover, in the case of the \textsf{Wikipedia2Vec} approach, it is not guaranteed that a large corpus should be more difficult than a small one, since the episodes are shorter for small corpora and therefore the algorithm has fewer steps to grasp the geometrical structures in the distribution of \textsf{Wikipedia2Vec} embeddings.

The diameter of the graph may also impact the performance of the model.
Indeed, Corpus 2 is again the one with the smallest diameter, which may have helped the model to determine the relationships between documents and keywords more quickly.
However, this must be balanced with the results on Corpus 6, on which our model performed far worse (in terms of normalized return) despite equal diameter.

Another noticeable result is the one of Corpus 5.
We remind that this corpus was the only one to be disconnected.
Actually, it was disconnected at the 11\textsuperscript{th} document, which is consistent with the performance of the model: indeed, episodic return lower than $10$ indicates that it failed to make recommendations beyond the 10\textsuperscript{th} document.
This can be explained quite simply: since this document is disconnected from the rest of the graph, it does not benefit from message-passing and therefore receives no information about other documents feedback.

Eventually, one cannot ignore the extremely high variance of the episodic return for almost all corpora (except for Corpus 2).
This is partly due to the choice of the \texttt{REINFORCE} algorithm, which is known for its high instability.

\subsubsection*{Ablation study}

We conducted an ablation study to analyse the contribution of \textsf{Wikipedia2Vec} embeddings compared to simple one-hot encodings.
Even though the approach with embeddings performed significantly better on each corpus, the high error margins and the similarity between trends suggest that our model was not truly able to leverage high level information about the relationships between Wikipedia entities.
Instead, it is more likely that it simply \say{overfit} to each corpus.
This lack of generalization is not a problem in the setting of our experiment but can be a serious issue in transfer learning scenarios and therefore needs to be addressed.

\section{Limitations and future work} \label{section:future_works}

\subsection{Size and structure of the graph}
 All of our experiments have been conducted on small graphs (less than $\sim 100$ nodes).
However, it is likely that our model would struggle a little more on larger graphs as the receptive field of each node accounts for a smaller fraction of the graph in such case.
Besides, it is not possible to increase the depth of a GNN indefinitely because of the over-smoothing problem \cite{kipf2016semi, wu2020comprehensive, li2018deeper}.
Therefore, it is likely that these embeddings alone would not be sufficiently informative to allow for long-term planning.
This limitation can be addressed with down- and upsampling methods such as pooling and unpooling operations on graphs, which make it possible to process information at multiple scales \cite{defferrard2016convolutional, simonovsky2017dynamic, ying2018hierarchical}.
It can also be addressed with planning techniques such as Monte Carlo Tree Search, which has demonstrated great performance in combination with deep RL techniques \cite{schrittwieser2020mastering, silver2016mastering, silver2010monte}.

As we saw in subsection \ref{section:results}, there is also an issue with disconnected graphs since our model failed to make predictions beyond the disconnected document node in Corpus 5.
One possible solution could be to slightly modify the structure of the graph, for example through link prediction based on keyword embeddings.

Eventually, it is important to note that we tested our approach on corpora related to engineering topics --- machine learning and programming ---  which keyword distributions might be quite similar (cf. Figure \ref{fig:keywords_evolution} in the Appendix).
Yet, corpora related to different topics may have completely different keyword distributions.
Therefore, it would be worth comparing the performance of the model on a wider range of subjects in the future.

\subsection{Variance and sample efficiency}

As stated in Section \ref{section:experiments}, our approach suffers from high variance, partly due to the choice of the \texttt{REINFORCE} algorithm.
Some other on-policy methods have demonstrated great success in reducing variance \cite{schulman2015trust, schulman2017proximal, mnih2016asynchronous}.
Nevertheless, these approaches remain generally not very sample-efficient.
To improve sample-efficiency, it is quite common to use off-policy algorithms as they allow to reuse past experience \cite{mnih2015human, lillicrap2015continuous, haarnoja2018soft}.
However, as stated in Section \ref{section:rl_agent}, the implementation of an approximate Q-value function with a GNN is not trivial as it requires to leverage information at the scale of the entire graph, which involves modifications in the model.
Another alternative is to use a model-based reinforcement learning algorithm (MBRL) \cite{sutton1990integrated, schrittwieser2020mastering}.
As they allow to learn a model of the environment (i.e. a model that predicts the next observations and rewards), MBRL techniques enable to reuse past experience and learn from a richer signal than the reward signal alone.
Therefore, they are usually much more sample-efficient than model-free RL techniques.
These approaches might be more appropriate in our case, as a local model like a GNN may more easily predict immediate feedback than the (long-term) \textit{value} of a state-action pair.

\subsection{Interpretability}

One of the main limitations of our approach is its lack of interpretability.
Ideally, an ITS would not only provide a personalized learning experience but also inform the learner about their progress and level of understanding, in order to encourage self-awareness and self-regulation.
This is usually done with an \textit{open learner model}.
However, like most deep learning approaches, our recommender system is a black-box model and does not allow for easy interpretation.
Yet, we hypothesize that the estimated knowledge state $\hat{\mathbf{s}}_t$ does not only contain semantic information about keywords but also about the way they were understood by the learner.
Therefore, future work may consist in projecting these keyword embeddings into lower dimensional space to visualize their evolution throughout learning sessions.


\subsection{Reusability}

We designed a model that is flexible enough to be \textit{theoretically} capable of transferring its knowledge from one corpus to another.
However, this is only possible if the model has managed to capture high-level information that is common to all corpora.
Unfortunately, our experiments do not allow to truly evaluate the transfer learning capabilities of our model.
However, since it seems to overfit to the structure of each corpus, it might not have learned that much about the high-level relationships in the distribution of \textsf{Wikipedia2Vec} embeddings.
Therefore, transfer learning might not be very effective in this case.
Future directions to reduce overfitting may consist in applying regularization techniques to GNN (such as node dropout), or using training techniques that push the model to learn higher-level knowledge, such as meta-learning for RL \cite{finn2017model}.

\section{Conclusion}\label{section:conclusion}

In this paper, we presented a new model for learning path personalization, designed to be reusable and independent of any expert labeling.
We demonstrated its ability to learn to make recommendations in 6 semi-synthetic environments made-up of real-world educational resources and simulated learners.
Since this model is theoretically capable of transferring its knowledge from one corpus to another, it is a first step towards an approach that could considerably reduce the cold-start problem.
Future work will investigate its performance in the context of transfer learning and with real students.

\section{Acknowledgments}
We would like to warmly thank Vincent François-Lavet from VU Amsterdam who gave us some great advice on the formalization of the reinforcement learning problem.
We would also like to thank Nicolas Vayatis, Argyris Kalogeratos (Centre Borelli), Nathanaël Beau and Antoine Saillenfest (onepoint) for their insightful reviews of the paper.

%
\bibliographystyle{abbrv}
\bibliography{rebibered}  
%

\pagebreak

\appendix



\begin{figure}[!h]
    \centering
    \includegraphics[width=\linewidth]{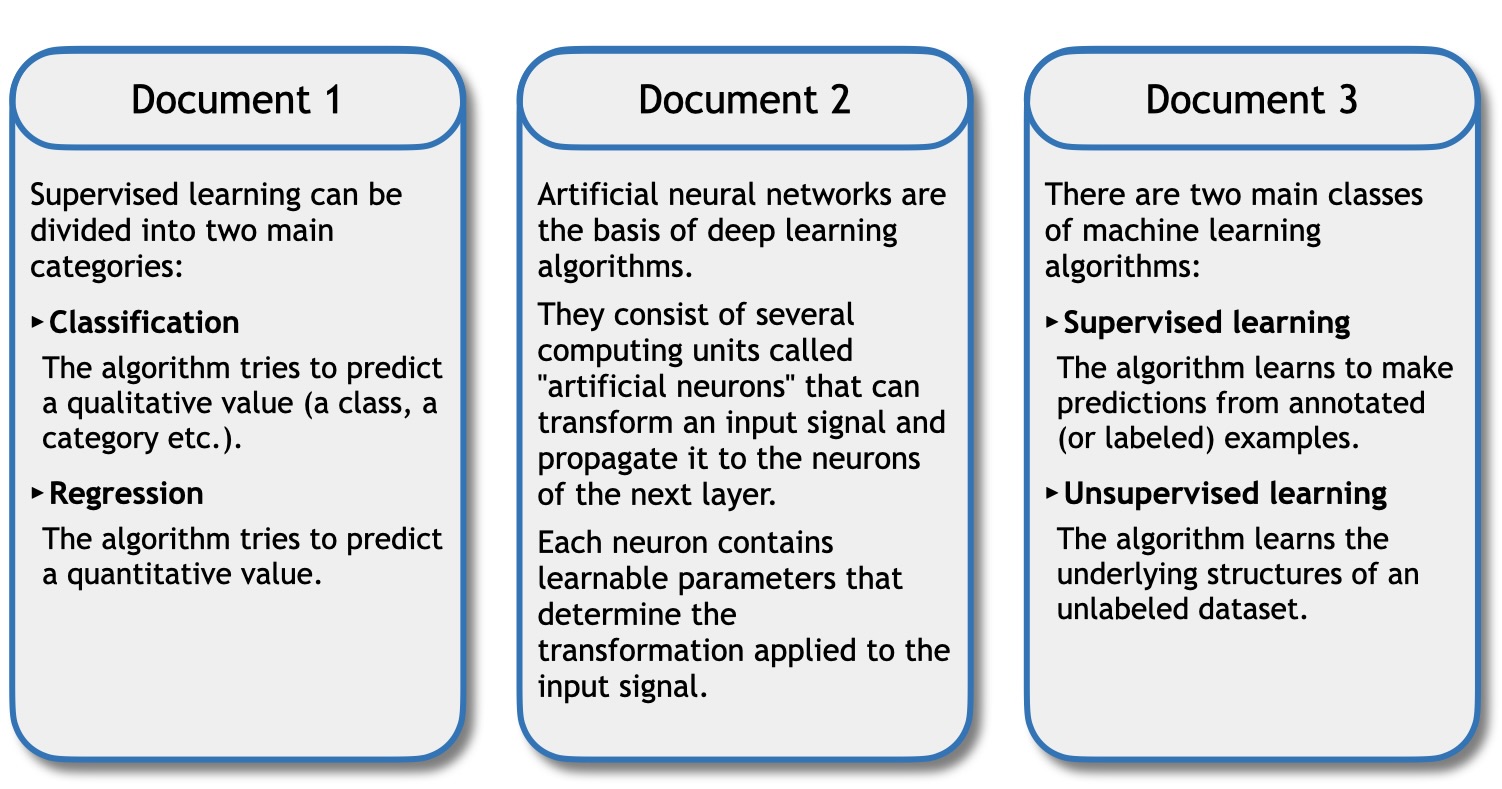}
    \caption{Three examples of \textit{self-contained} educational resources taken from a corpus dealing with machine learning basics}
    \label{fig:resources}
    \Description{3 documents containing keywords such as classification, regression, supervised learning, neural networks}
\end{figure}

\paragraph{Corpus and keywords}
Some examples of educational resources that satisfy the assumptions $(a_4)$, $(a_5)$ and $(a_6)$ described in Section~\ref{section:description} are provided in Figure \ref{fig:resources}.
In the document 1, an appropriate collection of keywords would be: \{\textit{supervised learning}, \textit{classification}, \textit{regression}\}.




\paragraph{Linear corpus}

\begin{figure}[h]
    \centering
    \includegraphics[width=\linewidth]{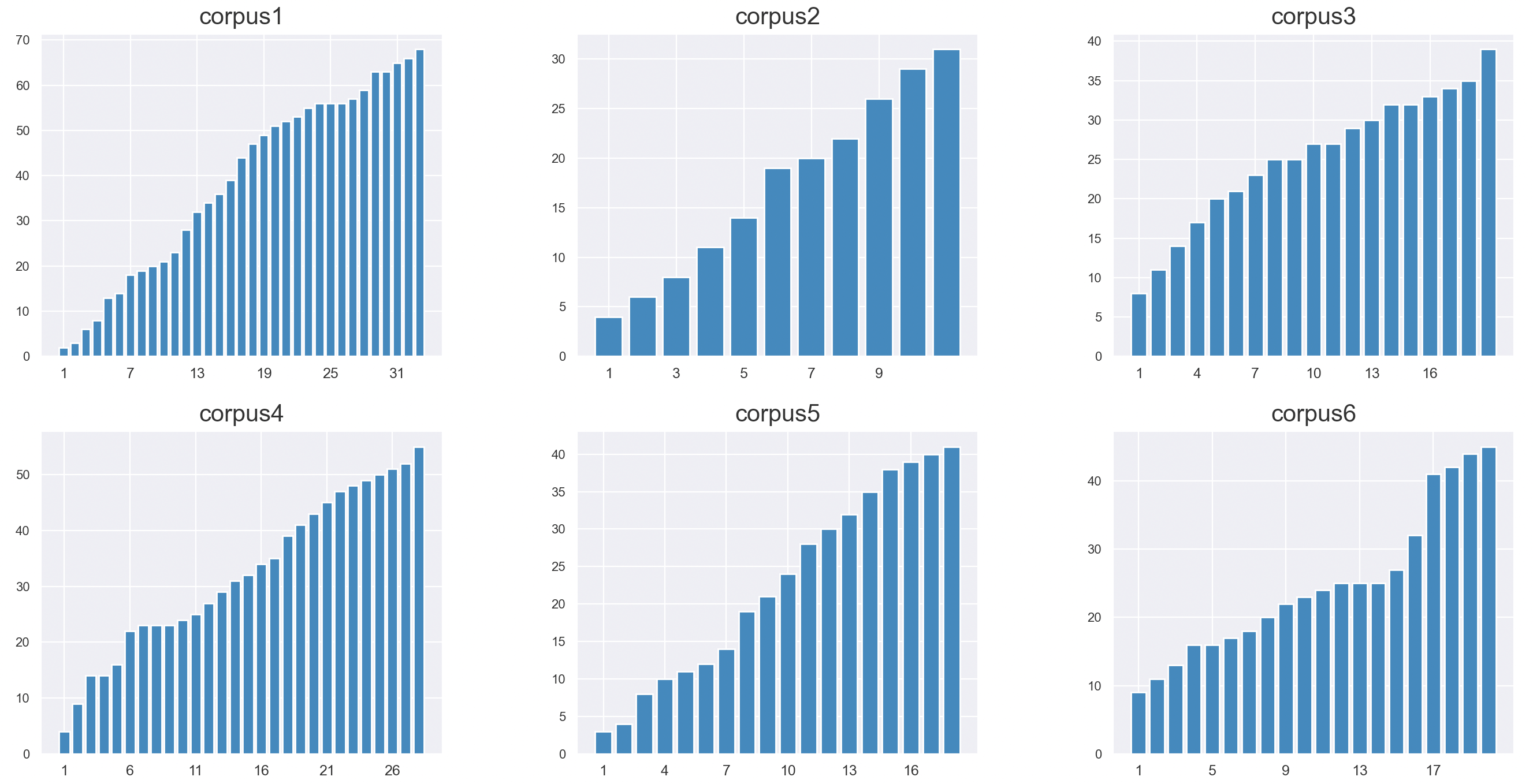}
    \caption{Evolution of the total number keywords in each course}
    \label{fig:keywords_evolution}
    \Description{Histogram representing the number of new keywords brought by each document along the course}
\end{figure}

In our experiments, we used 6 corpora\break based on courses taken from a popular \textit{e-learning} platform.
Figure \ref{fig:keywords_evolution} shows the evolution of the total number of keywords throughout each course.
Note that although they all cover different topics and were designed by different educators, they always introduce new keywords in a \say{linear} way.
This supports the idea that the distribution of keywords can be a good indicator of pre-requisite relationships between documents.

\paragraph{Simulated learners}
In the following we present a step by step example of a learning path followed by a simulated learner (also detailed in Table~\ref{tab:simulated-students}).

Consider a corpus of three documents $\{d_1, d_2, d_3\}$, designed to be explored in the order of indices: $d_1$ is a prerequisite for $d_2$ and $d_2$ is a prerequisite for $d_3$.
A simulated student can understand a document only if they have understood its prerequisites.
Throughout the learning path, we maintain a set $\mathcal{D}_{\circ}$ of understood documents, initialized as an empty set: $\mathcal{D}_{\circ} = \{\}$. 

\begin{table}[t]
    \centering
    \begin{tabular}{ccccc} \toprule
        step & action $\mathbf{a}_t $ & feedback $\mathbf{f}_t$ & reward $\mathbf{r}_t$ & $\mathcal{D}_\circ$\\ \midrule
        1 & $d_2$ & $f_<$  & 0 & $\{\}$\\
        2 & $d_1$ & $\feq$ & 1 & $\{d_1\}$\\
        3 & $d_3$ & $f_<$  & 0 & $\{d_1\}$\\
        4 & $d_2$ & $\feq$ & 1 & $\{d_1, d_2\}$\\
        5 & $d_1$ & $f_>$  & 0 & $\{d_1, d_2\}$\\
        6 & $d_3$ & $\feq$ & 1 & $\{d_1, d_2, d_3\}$\\ \bottomrule
    \end{tabular}
    \caption{Example of a sequence of interactions (learning path) between a simulated student and our policy}
    \label{tab:simulated-students}
\end{table}

At step 1, the policy recommends document $d_2$ (with prerequisite $d_1$). $d_1\notin\mathcal{D}_{\circ}$, therefore the student returns feedback $(f_{<})$. At step 2, the policy recommends document $d_1$. This document has no prerequisite, therefore the student returns feedback $(\feq)$ and we add $d_1$ to $\mathcal{D}_{\circ}$. At step 3, the policy recommends document $d_3$. $d_2\notin\mathcal{D}_{\circ}$, therefore the student returns feedback $(f_{<})$. At step 4, the policy recommends document $d_2$. $d_1\in\mathcal{D}_{\circ}$, therefore the student returns feedback $(\feq)$ and we add $d_2$ to $\mathcal{D}_{\circ}$. At step 5, the policy recommends document $d_1$. $d_1\in\mathcal{D}_{\circ}$, therefore the student returns feedback $(f_{>})$. Finally at step 6, the policy recommends document $d_3$. $d_2\in\mathcal{D}_{\circ}$, therefore the student returns feedback $(\feq)$ and we add $d_3$ to $\mathcal{D}_{\circ}$.

Note that in this example, we fixed $T=6$ to display a greater number of situations.
Conversely, in our experiments, $T$ was always equal to the size of the corpus.




\paragraph{Self-attention}

The self-attention coefficient $\alpha_{w d}$ used in Equations \ref{eq:kw2doc} and \ref{eq:doc2kw} is defined as follows.
For any nodes $w$, $d$:
\begin{equation}
    \alpha_{w d}^{(\ell)}=\underset{\mathcal{N}(w)}{\operatorname{softmax}}\left( a\left(W^{(\ell)} \mathbf{h}_d^{(\ell)}, W^{(\ell)} \mathbf{h}_w^{(\ell)}\right) \right)
\end{equation}
where $W^{(\ell)}\in\mathbb{R}^{K\times K}$ refers to the weights of $\ell$th layer and \mbox{$a: \mathbb{R}^{K} \times \mathbb{R}^{K} \rightarrow \mathbb{R}$} is the additive attention mechanism.
The $\operatorname{softmax}$ function is taken over all neighbors of node $w$ (further details below).

\paragraph{Multilayer perceptron}

Each $\textsf{MLP}_{K_1 \to K_2}(\cdot)$ operator used in Section \ref{section:rl_agent} is a multilayer perceptron with one hidden layer.
For any input vector $\mathbf{x}$, this operation boils down to:
\begin{equation}
    \mathbf{x}^\prime = A^{(2)} \sigma \left(A^{(1)} {\mathbf{x}} + B^{(1)} \right) + B^{(2)}
\end{equation}
where $\sigma(\cdot)$ is the ReLU activation function, $A^{(1)} \in \mathbb{R}^{K_1\times K}$, $A^{(2)} \in \mathbb{R}^{K\times K_2}$, $B^{(1)} \in \mathbb{R}^{K}$ and $B^{(2)} \in \mathbb{R}^{K_2}$ are trainable parameters.

\paragraph{Softmax operator}

The softmax operator over a finite collection $E$ of real numbers is defined as follows:
\begin{equation} 
    \forall x \in E, \quad \underset{E}{\operatorname{softmax}}({x}) = \frac{\exp{x}}{\sum_{y\in E} \exp{y}}.
\end{equation}


\begin{figure*}[ht]
    \centering
    \includegraphics[width=\linewidth]{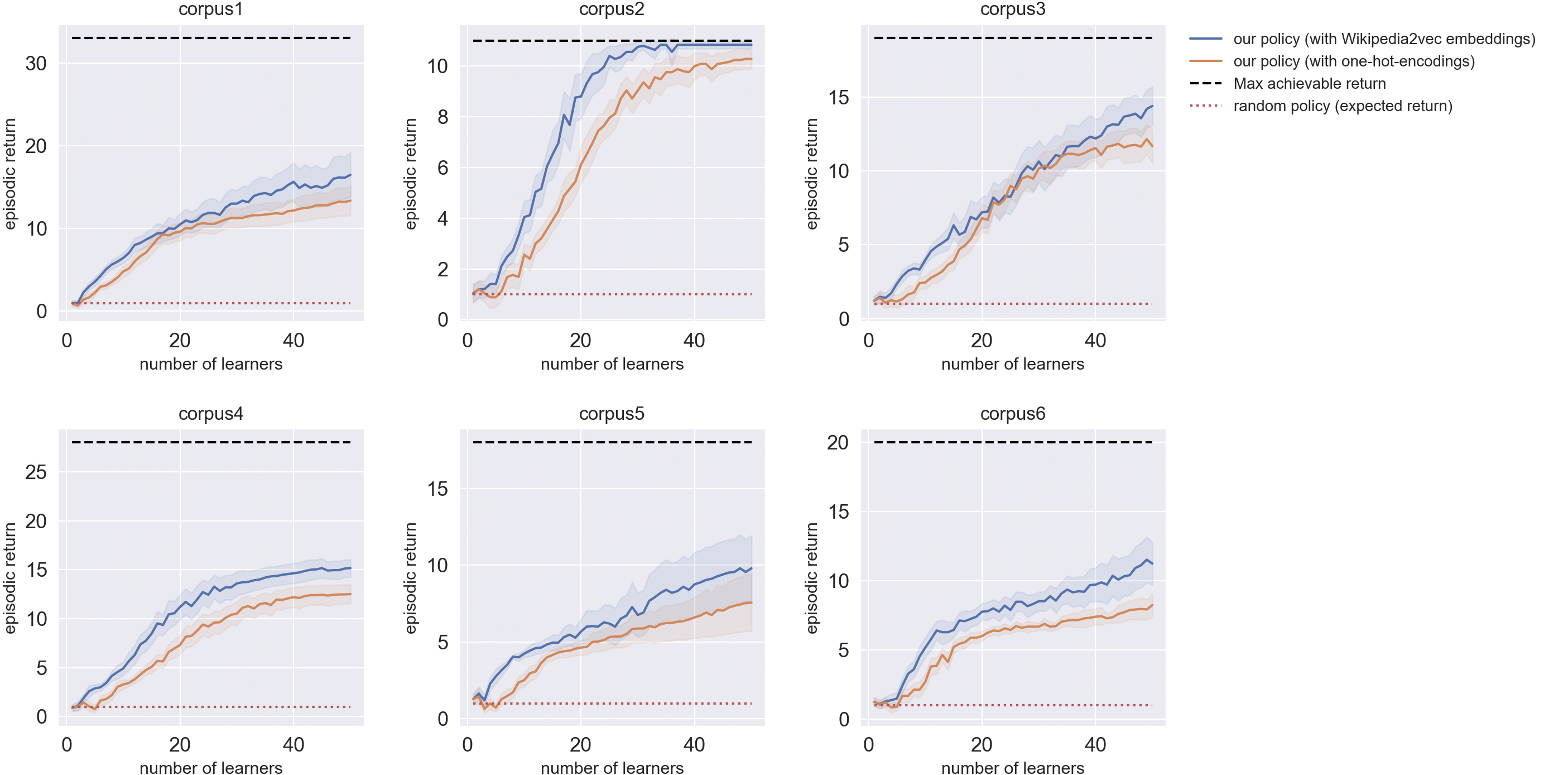}
    \caption{Evolution of the episodic return on 50 simulated learners for 6 corpora}
    \label{fig:results_curves}
    \Description{Return increases after each episode of learning, for our policy compared to a random policy}
\end{figure*}

\paragraph{Wikipedia2Vec}

The pretrained \textsf{Wikipedia2Vec} embeddings leveraged as keyword features in our experiment were derived from a skip-gram model trained on a triple objective: (1) predicting neighboring entities in the link graph of Wikipedia, (2) predicting neighboring words given each word in a text contained on a Wikipedia page, and (3) predicting neighboring words given a target entity using anchors and their context words in Wikipedia \cite{wikipedia2vec}.
We hypothesize that in addition to modeling the semantic information carried by each keyword, these embeddings allow to capture prerequisite relationships between concepts, especially through task (1).

\paragraph{Experimental results}
The learning curves of our experiments are reported in Figure \ref{fig:results_curves}.
For reproducibility, we also reported the hyperparameters of our model in Table \ref{table:hyperparameters}.

\begin{table}[ht]
    \centering
    \medskip
    \begin{tabular}{ccc}
        \toprule
        Name & Value\\ \midrule
        Learning rate                          & $0.0005$  \\
        Hidden dimension                       & $32$  \\
        Activation function                    & ReLU   \\
        Attention type                         & additive   \\
        Number of attention heads              &  $2$   \\
        \textsf{Wikipedia2Vec} embedding size & $100$ \\
        Documents encoding                     &  vector of zero\\
        Feedback encoding                      &  one-hot-encoding  \\
        Remaining time encoding                &  counter  \\
        Batch size                             &  16  \\
        Repeat per collect                     &  15  \\
        Episodes per collect                   &  1  \\
        \bottomrule
    \end{tabular}
    \caption{Hyperparameters used in our policy model}
    \label{table:hyperparameters}
\end{table}

\end{document}